\documentclass{article}
\usepackage{spconf,amsmath,epsfig}
\usepackage{algorithmic}
\usepackage{graphicx}
\usepackage{textcomp}
\usepackage{xcolor}
\usepackage{subfigure}
\usepackage{cite}
\usepackage{amssymb}
\usepackage{multirow}

\title{ReAFFPN:Rotation-equivariant Attention Feature Fusion Pyramid Networks for Aerial Object Detection}
\name{Chongyu Sun, Yang Xu, Zebin Wu, Zhihui Wei\thanks{ This work was supported in part by the National Natural Science Foundation of China (61772274, 62071233, 61971223, 61976117), the Jiangsu Provincial Natural Science Foundation of China (BK20211570, BK20180018, BK20191409), the Fundamental Research Funds for the Central Universities (30917015104, 30919011103, 30919011402, 30921011209), and in part by the China Postdoctoral Science Foundation under Grant 2017M611814, 2018T110502. }}
\address{School of Computer Science and Enginnering, Nanjing University of Science and Technology, Nanjing, \\Jiangsu Province, 210094, China}
\begin{document}
%
\maketitle
\begin{abstract}
This paper proposes a Rotation-equivariant Attention Feature Fusion Pyramid Networks for Aerial Object Detection named ReAFFPN. ReAFFPN aims at improving the effect of rotation-equivariant features fusion between adjacent layers which suffers from the semantic and scale discontinuity. Due to the particularity of rotational equivariant convolution, general methods are unable to achieve their original effect while ensuring rotation equivariance of the network. To solve this problem, we design a new Rotation-equivariant Channel Attention which has the ability to both generate channel attention and keep rotation equivariance. Then we embed a new channel attention function into Iterative Attentional Feature Fusion (iAFF) module to realize Rotation-equivariant Attention Feature Fusion. Experimental results demonstrate that ReAFFPN achieves a better rotation-equivariant feature fusion ability and significantly improve the accuracy of the Rotation-equivariant Convolutional Networks.
\end{abstract}
\begin{keywords}
Object Detection, Remote Sensing, Rotation Equivariance, Attention, Feature Fusion
\end{keywords}

\begin{figure*}[!t]
	\centering
	\subfigure[ReAFFPN] {\includegraphics[width=0.6\textwidth]{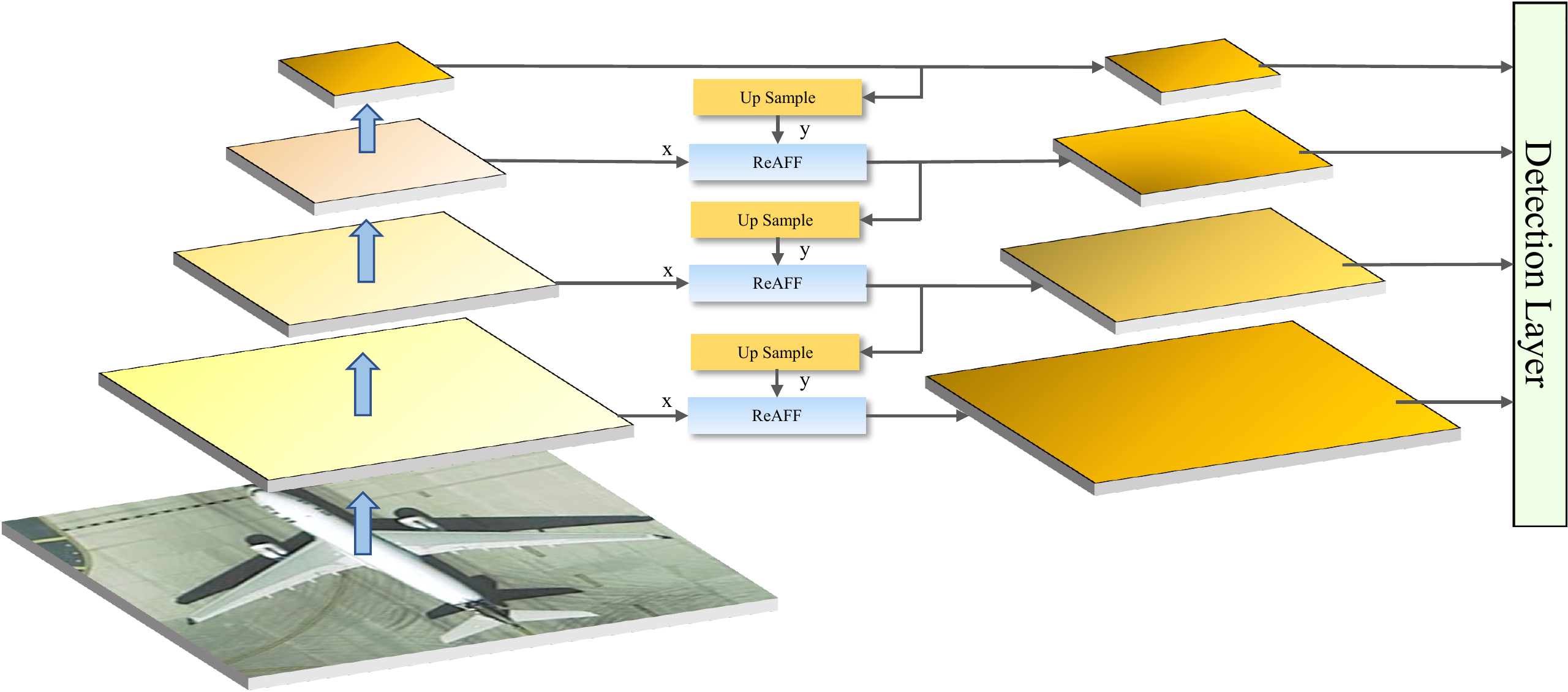}}
	\subfigure[Rotation-equivariant feature maps] {\includegraphics[height=0.2\textheight]{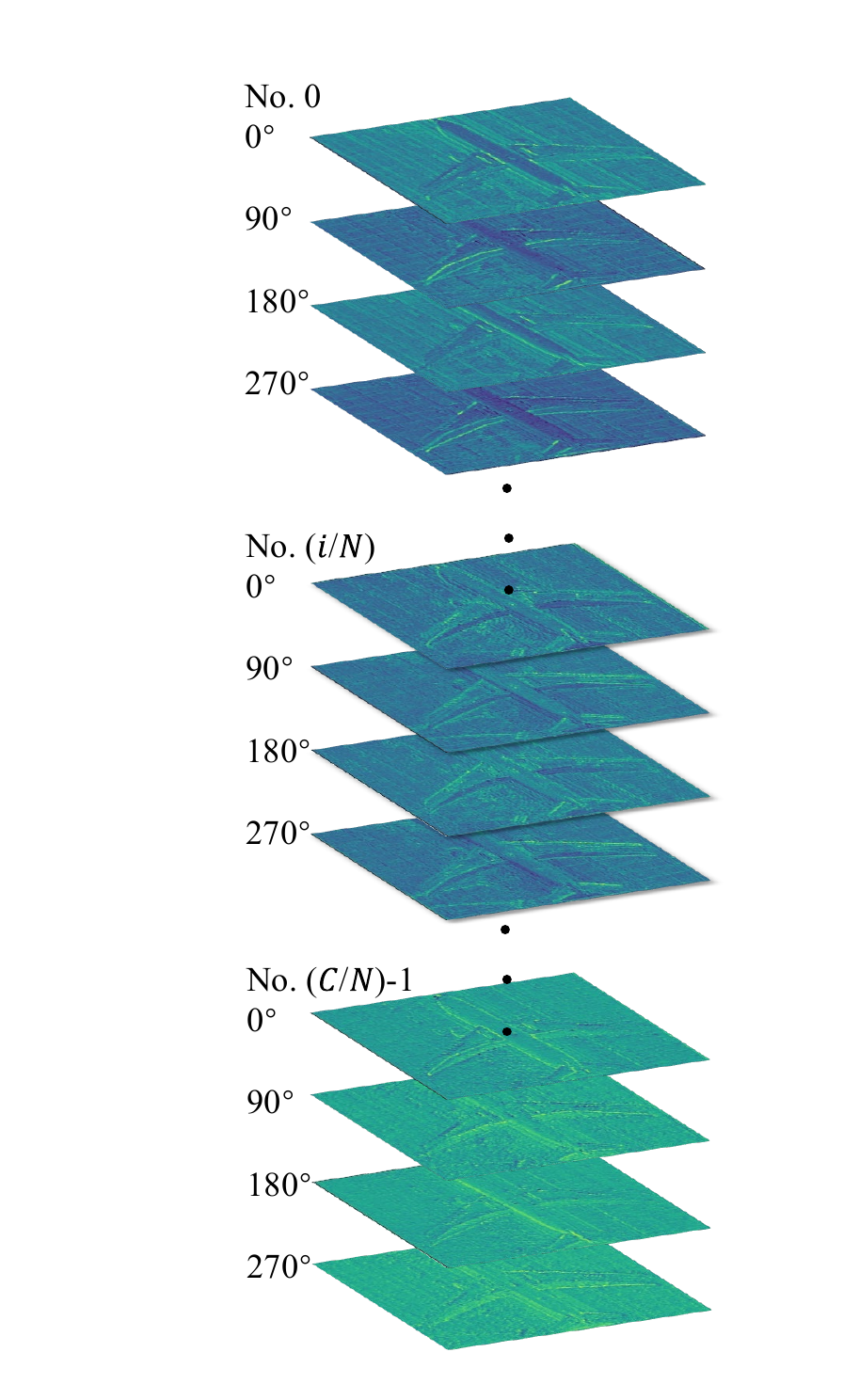}}
	\caption{ Overview of our proposed method. (a) is the basic structure of ReAFFPN. (b) is the architecture of re-feature maps in ReCNNs. $N$ is the orientation parameter which is set to be 4 for clarity, $C$ is the total channel number, and $r$ is reduction ratio. The channels of re-feature maps have two parts, rotation-equivariant convolution kernel channels (we will use kernel channels in the following part for simplicity) and orientation channels, the kernel channels are independent of each other while orientation channels of same group convolutional kernel have similar semantics information but different orientation information. }
	\label{fig5}
\end{figure*}

\section{Introduction}
\label{sec:intro}

Object detection based on deep neural network in aerial images has become increasingly significant in recent years. Different from those in natural images, objects in aerial images are often distributed with arbitrary orientation and different scales, which makes detecting aerial objects more challenging. To better represent objects with arbitrary orientation, most aerial detection methods used Oriented Bounding Boxes (OBBs) instead of Horizontal Bounding Boxes (HBBs). Meanwhile, Feature Pyramid Network (FPN) is used to cope with the scale variation problem. 

The key to object detection is feature extraction, which demands that networks should extract both equivariant features for localization and invariant features for classification. For natural images detection task, ordinary CNNs are capable of extracting equivariant features because most objects only have translation and scale changes, then with RoI Align or other operations, invariant features will be generated. In aerial images, however, ordinary CNNs cannot generate totally equivariant features due to its lacking of rotation-equivariance. Cohen et al. [1] proposed Group Equivariant Convolutional Networks(G-CNNs), a new convolution form that enables networks to formulate orientation information, which uses group convolution to extract rotation-equivariant features. Weiler et al. [2] designed Steerable Filter CNNs (SFCNNs) based on G-CNNs to avoid interpolation artifacts from filter rotation. In aerial image field, rotation equivariance is firstly introduced into arbitrary orientation object detection in ReDet [3], in which convolutions of backbone and FPN are all rotation equivariant forms. With the translation and rotation equivariance of the convolution kernel, feature maps in ReDet provides more precise position and orientation information. With RoI Transformer [4] and RiRoI Align, ReDet is able to generate translation and rotation invariant features. However, due to the Rotation-equivariant Convolution Networks’ (ReCNNs) introducing additional orientation channels, feature maps in ReCNNs have less semantic information compared with that of ordinary CNNs in the condition of same channel numbers, which will affact the classification accuracy. A feasible way to make networks better use extracted semantic features is channel attention enhancement, different from normal types, the iAFF [5] used both global and local channel attention to avoid the disappearance of small objects' feature response. However, without considering the orientation channels, existing channel attention mechanism will destroy rotation equivariance. Even worse, the problem of lacking semantic information will aggravate the semantic inconsistency between different feature layers in FPN, which limits the effect of feature fusion in top-down ways.

To address the above issues, we propose a Rotation-equivariant Attention Feature Fusion Pyramid Networks (ReAFFPN) for Aerial Object Detection, which aims at solving the semantic insufficiency problem and feature fusion semantic inconsistency problem of ReCNNs.

1.We design a new channel attention form called Rotation-equivariant Channel Attention. The proposed ReCA has the ability to enhance rotation-equivariant feature maps’ semantic information while keeping rotation-equivariance in ReCNNs.

2.We improve the Iterative Attentional Feature Fusion Module (iAFF) [5] with our proposed Rotation-equivariant Channel Attention called ReAFF and design a Rotation-equivariant Attention Feature Fusion Pyramid Networks (ReAFFPN) for better rotation-equivariant feature fusion in aerial object detection.

\section{Proposed Method}
\label{sec:Proposed Method}
The details of our proposed Rotation-equivariant Attention Feature Fusion Pyramid Networks (ReAFFPN) are presented in this section, the basic structure is shown in Fig.1 (a). First, we design a new Rotation-equivariant Channel Attention (ReCA) with the ability of keeping rotation-equivariance and enhancing channel attention at the same time. And then, we improve the iAFF with our ReCA, and design a ReAFFPN for better a equivariant feature maps fusion.

\subsection{Rotation-equivariant Channel Attention}
The current channel attention forms are all designed for ordinary CNNs, which deem all channels as independent to generate channel weights. In ReCNNs, however, channels are not completely unrelated. As shown in Fig.1 (b), the channels of rotation-equivariant feature maps (in the following we will use re-feature maps for simplicity) have two parts, the group convolutional kernel channels and orientation channels. If we directly apply the ordinary channel attention to the re-feature maps, the rotation information of orientation channels will be ignored, which will ruin the equivariance of re-feature maps.

\begin{figure*}[!t]
	\centering
	\subfigure[ReCA] {\includegraphics[width=0.6\textwidth]{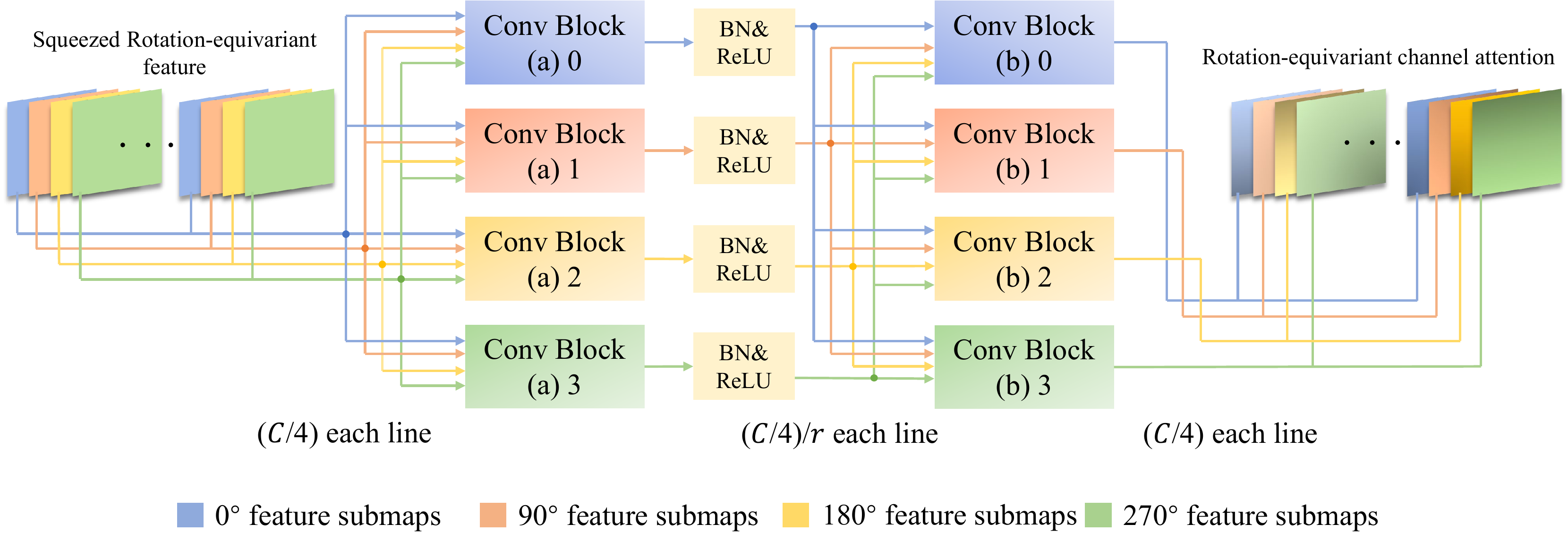}}
	\subfigure[Conv Blocks] {\includegraphics[height=0.18\textheight]{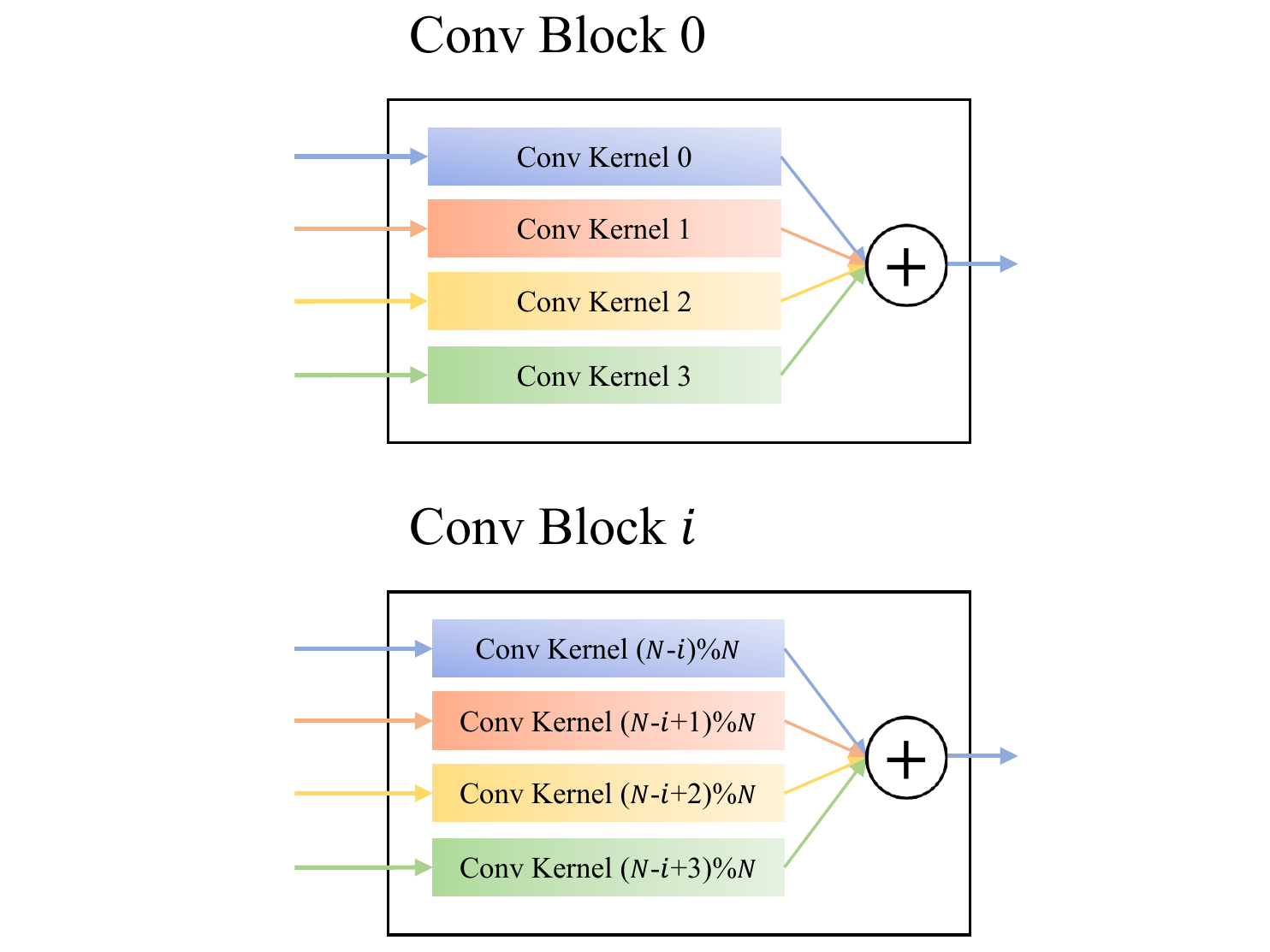}}
	\caption{ Overview of the ReCA. (a) is the overall process of Rotation-equivariant Channel Attention, each line represents squeezed re-feature submaps with the same orientation information, and the channel number of submaps is marked at the bottom of (a). (b) is the inner structure of the Conv Blocks, each Conv Kernel takes a re-feature submap as input and takes the result of channel wise addition as output. Conv Kernels with the same id share their weights, and kernel positions’ cyclic shifts are implemented among different Conv Blocks, $N$ is the orientation paramater which is set to be 4 for clarity. }
	\label{fig5}
\end{figure*}

In order to keep rotation-equivariance while enhancing the channel attention of re-feature maps, we proposed the ReCA. The structure of ReCA is shown in Fig.2 (a), the ReCA takes squeezed re-feature maps (squeeze operation is spatial dimension global average pooling in our experiments) as input. Then, the input will firstly be divided into $N$ (orientation parameter in rotation-equivariant networks) re-feature submaps according to the orientation channels, each submap contains an orientation information. Then re-feature submaps will go through $N$ Conv Blocks a, Batch Normalization layers, ReLU layers, $N$ Conv Blocks b and a sigmoid function to extract rotation-equivariant attention weights, the weights will be rearranged as the original re-feature maps. Fig.2 (b) shows the inner structure of Conv Block which has $N$ Conv Kernels. In Conv Block (a), the input size of Conv Kernel is $1\times1\times(\frac{C}{N})$ and output size is $1\times1\times(\frac{C}{N})/r$, and vice versa in Conv Block (b). The output of each Conv Kernels will go through a channel wise addition operation to generate the result of a Conv Block, which can be represented as follows:

\begin{equation}\label{eq}
	\begin{split}
		& CB_{i}^a=M_{a}(F_{squ})=\sum_{n=0}^{N-1} W_{(n-i)\%N}^a(F_{squ(n)}),\\
	    & CB_{i}^b=M_{b}(CB^a)=\sum_{n=0}^{N-1} W_{(n-i)\%N}^a(CB_{n}^a),
	\end{split}
\end{equation}
where $F_{squ}$ is squeezed re-feature maps and $F_{squ(n)}$ is squeezed re-feature submaps of orientation $n$. $CB^a$ and $CB^b$ represent the output of $N$ Conv Blocks (a) and (b) whose channel wise are distributed as $F_{squ}$. $W_{n}^a\in\mathbb{R}^{(\frac{C}{N})\times(\frac{C}{N})/r}$ and $W_{n}^b\in\mathbb{R}^{(\frac{C}{N})/r\times(\frac{C}{N})}$ represent the weights of Conv Kernels in Conv Block (a) and (b).

In order to fully use the orientation information, weight sharing and cyclic shift is adopted in Re-ChAtt, through which the channel attention weights with $N$ orientational information will be generated. The whole process can be formulated as follows:
\begin{equation}
	ReM_{c}(F)=\sigma(CB^b),\label{eq}
\end{equation}
where $\sigma$ is sigmoid function.

We embed the Rotation-equivariant Channel Attention into Rotation-equivariant Feature Pyramid Network (ReFPN) to enhance the channel attention of the upper features before each top-down path. The effectiveness of our ReCA is demonstrated through comparative experiments. The details are in Table 1.

\begin{figure}[!t]
	\centering
	\subfigure[ReAFF] {\includegraphics[width=0.2\textwidth]{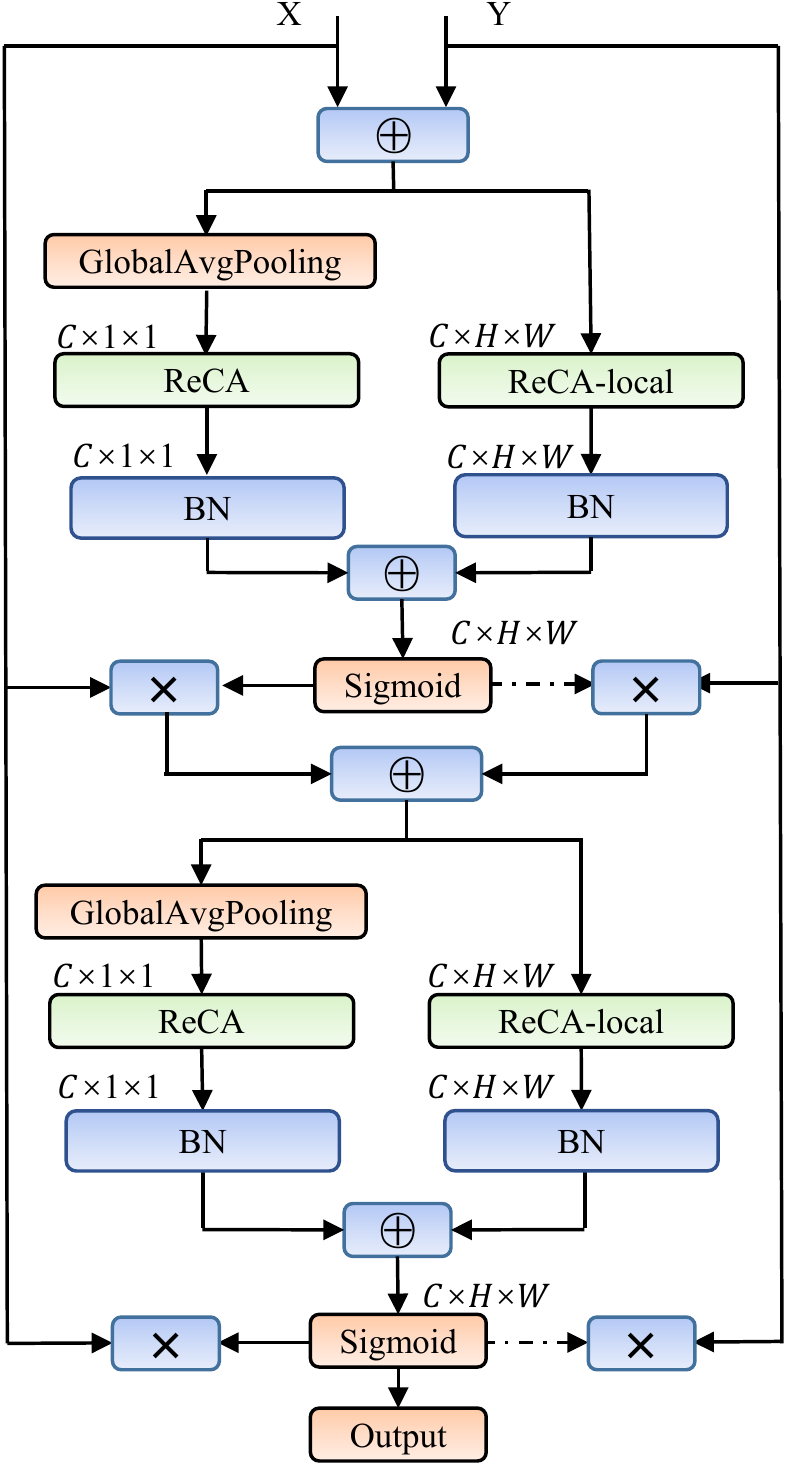}}
	\caption{ The basic structure of ReAFF.  }
	\label{fig5}
\end{figure}

\subsection{Rotation-equivariant Attention Feature Fusion Pyramid Networks}
Though ReCA has the ability to relieve the semantic insufficient problem, however, the ReCNNs still suffer from the scale and semantic discontinuity among different feature layers in ReFPN. For the purpose of better fusing re-feature maps, we introduce iAFF and improve it with our ReCA, named ReAFF. The ReAFFPN are designed by embedding ReAFF into ReFPN.

The structure of ReAFF with two attention feature fusion parts is shown in Fig.3. The first part is designed for better initially integrating input features while the second part is designed for the final attention feature fusion. Two channel attention strategies are used. In addition to the general channel attention on the left branch in Fig.3, the local channel attention is embedded on the right branch which will retain the channel signal of small objects. The original iAFF cannot directly be used in ReCNNs for the reason its channel attention will damage the rotation-equivariance. As a result, we replace the channel attention function with our ReCA to make this module rotation-equivariant. Then, we embedded it into ReFPN as presented in Fig.1 (a), each upper re-feature maps will go through an up-sample operation then fuse with a lower re-feature map through ReAFF. 

With ReCA and ReAFF, the feature fusion in ReCNNs will be carried out with less scale and semantic discontinuity. Our experiments show the effectives of our proposed ReAFFPN. The details are shown in Table 1 and Table 2.

\section{Experiments}
\label{sec:Experiments}

\subsection{Experimental Details}
\textbf{DOTA}[6] is the largest dataset used to evaluate performance of orientated aerial object detection. \textbf{DOTA} has two version: \textbf{DOTA-v1.0} and \textbf{DOTA-v1.5} which contain 2806 large aerial images with size from 800×800 to 4000×4000. The differences between two version are that \textbf{DOTA-v1.0} has 188,282 instances among 15 categories: Plane (PL), Baseball diamond (BD), Bridge (BR), Ground track field (GTF), Small vehicle (SV), Large vehicle (LV), Ship (SH), Tennis court (TC), Basketball court (BC), Storage tank (ST), Soccer-ball field (SBF), Roundabout (RA), Harbor (HA), Swimming pool (SP), and Helicopter (HC), while \textbf{DOTA-v1.5} containing 402,089 instances has a new category, Container Crane (CC) and many small instances with size less than 10 pixels.

\textbf{HRSC2016}[7] is another orientated aerial object dataset with a ship class which contains 1061 images with size ranging from 300×300 to 1500×900.

Our experiment settings are consistent with those in ReDet during training and testing. ReCNNs with RiRoiAlign is adopted as baseline method. We used 2 NVIDIA RTX 3090 GPUs for training and testing, and we only trained and tested in single scale with no other data augmentation, therefore we only compare with the non-data-augmentation results of the baseline method.

\subsection{Experimental Results}
The mean average precision of all classes (mAP) is adopted to evaluate the performance of the module on \textbf{DOTA}, while mAP, AP50 and AP75 are adopted on \textbf{HRSC2016}. 

As shown in Table 1, Channel Attention and iAFF have bad influence on the accuracy of ReCNNs on \textbf{DOTA-v1.0} for damaging the rotation-equivariance, while our proposed ReCA and ReAFFPN which retain the rotation-equivariance had improved the performance on \textbf{DOTA-v1.0} by \textbf{0.67} and \textbf{0.88} respectively. \textbf{DOTA-v1.5} includes many extremely tiny objects compared with \textbf{DOTA-v1.0}, which makes object detection more difficult, we evaluated our ReAFFPN on \textbf{DOTA-v1.5} and got the improvement of mAP by \textbf{0.32} compared with the baseline method.

\begin{table}[htbp]
	\caption{Performance comparison of different methods on \textbf{DOTA}.}
	\begin{tabular}{c|cl|cl}
		\hline
		\multicolumn{1}{l|}{DOTA version} & \multicolumn{2}{c|}{Method}                  & \multicolumn{2}{c}{mAP}          \\ \hline
		\multirow{5}{*}{DOTA-v1.0}        & \multicolumn{2}{c|}{baseline}                & \multicolumn{2}{c}{76.25}        \\
		& \multicolumn{2}{c|}{ReFPN+Channel Attention} & \multicolumn{2}{c}{71.99(-4.26)}             \\
		& \multicolumn{2}{c|}{ReFPN+ReCA}           & \multicolumn{2}{c}{76.92(+0.67)} \\
		& \multicolumn{2}{c|}{ReFPN+iAFF}              & \multicolumn{2}{c}{64.95(-11.3)} \\
		& \multicolumn{2}{c|}{ReAFFPN}                 & \multicolumn{2}{c}{\textbf{77.13(+0.88)}} \\ \hline
		\multirow{2}{*}{DOTA-v1.5}        & \multicolumn{2}{c|}{baseline}                & \multicolumn{2}{c}{66.86}        \\
		& \multicolumn{2}{c|}{ReAFFPN}                 & \multicolumn{2}{c}{\textbf{67.17(+0.31)}} \\ \hline
	\end{tabular}
\end{table}

HRSC2016 is an aerial dataset containing a large number of thin and long ship instances with arbitrary orientation. We compare our ReAFFPN with the baseline method in AP50, AP75 and mAP, and our method has improved the performance on \textbf{HRSC2016} by \textbf{0.1}, \textbf{0.24} and \textbf{1.59} respectively.

\begin{table}[htbp]
	\caption{Performance comparison of different methods on \textbf{HRSC2016}.}
	\begin{tabular}{l|cl|cl|cl}
		\hline
		Method   & \multicolumn{2}{c|}{AP50}        & \multicolumn{2}{c|}{AP75}         & \multicolumn{2}{c}{mAP}          \\ \hline
		baseline & \multicolumn{2}{c|}{90.46}       & \multicolumn{2}{c|}{89.46}        & \multicolumn{2}{c}{70.41}        \\
		ReAFFPN  & \multicolumn{2}{c|}{\textbf{90.56(+0.1)}} & \multicolumn{2}{c|}{\textbf{89.70(+0.24)}} & \multicolumn{2}{c}{\textbf{72.09(+1.59)}} \\ \hline
	\end{tabular}
\end{table}

\section{Conclusion}
\label{sec:conclusion}

This paper presents a Rotation-equivariant Attention Feature Fusion Pyramid Networks for Aerial Object Detection named ReAFFPN which aims at improving the performance of Rotation-equivariant CNNs while keeping the equivariance of the networks. With the proposed Rotation-equivariant Channel Attention, ReAFF module and ReAFFPN were designed to address the scale and semantic inconsistency problem between different rotation-equivariant feature layers. The Experimental results show that ReAFFPN can effectively improve the performance of ReCNNs.

\bibliographystyle{IEEEbib}
\bibliography{strings,refs}

\begin{thebibliography}{00}
\label{sec:ref}
	\bibitem{b1} T. Cohen and M. Welling, "Group equivariant convolutional networks", International Conference on Machine Learning (ICML), 2016.
	\bibitem{b2} M. Weiler, F. A. Hamprecht and M. Storath, "Learning Steerable Filters for Rotation Equivariant CNNs," 2018 IEEE/CVF Conference on Computer Vision and Pattern Recognition, 2018, pp. 849-858.
	\bibitem{b3} J. Han, J. Ding, N. Xue and G. -S. Xia, "ReDet: A Rotation-equivariant Detector for Aerial Object Detection," 2021 IEEE/CVF Conference on Computer Vision and Pattern Recognition (CVPR), 2021, pp. 2785-2794.
	\bibitem{b4} J. Ding, N. Xue, Y. Long, G. Xia and Q. Lu, "Learning RoI Transformer for Oriented Object Detection in Aerial Images," 2019 IEEE/CVF Conference on Computer Vision and Pattern Recognition (CVPR), 2019, pp. 2844-2853.
	\bibitem{b5} Y. Dai, F. Gieseke, S. Oehmcke, Y. Wu and K. Barnard, "Attentional Feature Fusion," 2021 IEEE Winter Conference on Applications of Computer Vision (WACV), 2021, pp. 3559-3568.
	\bibitem{b6} G.-S. Xia, X. Bai, J. Ding, Z. Zhu, S. Belongie, J. Luo, M. Datcu, M. Pelillo, and L. Zhang. DOTA: A large-scale dataset for object detection in aerial images. In CVPR, pages 3974-3983, 2018.
	\bibitem{b7} Z. Liu, L. Yuan, L. Weng, and Y. Yang. "A high resolution optical satellite image dataset for ship recognition and some new baselines," In ICPRAM, pages 324-331, 2017.
	
\end{thebibliography}

\end{document}